# Lexical Disambiguation in Natural Language Questions (NLQs)

Omar Al-Harbi[1] , Shaidah Jusoh[2,] and Norita Md Norwawi[3]

[1] **Faculty of Science and Technology, Islamic Science University of Malaysia, Malaysia**

[2] **Dept of Computer Science, Faculty of Science & Information Technology, Zarqa University, Zarqa, Jordan**

[3] **Faculty of Science and Technology, Islamic Science University of Malaysia, Malaysia**

## Abstract

Question processing is a fundamental step in a question answering (QA) application, and its quality impacts the performance of the QA application. The major challenging issue in processing question is how to extract semantic of natural language questions (NLQs). A human language is ambiguous. Ambiguity may occur at two levels; lexical and syntactic. In this paper, we propose a new approach for resolving lexical ambiguity problem by integrating context knowledge and concepts knowledge of a domain, into shallow natural language processing (SNLP) techniques. Concepts knowledge is modeled using ontology, while context knowledge is obtained from WordNet, and it is determined based on neighborhood words in a question. The approach will be applied to a university QA system.
***Keywords:*** *Question Answering (QA), Word Sense Disambiguation (WSD), Shallow Natural Language Processing (SNLP), WordNet, Context Knowledge, Ontology.*

## 1. Introduction

The QA area has attracted computational linguistics community in the last two decades. The aim of QA is to automatically return an exact answer to a NLQ instead of a list of documents. QA is composed of three types of processes: question processing, documents processing, and answer processing. A NLQ is the primary source through which a search process is directed for answers. Therefore, a careful and accurate analysis to the question is required. Thus, question processing is the most fundamental step in a QA application, and its quality impacts the performance of the QA application.

QA field is moving from only depending on retrieving and matching to understanding and reasoning natural language techniques [1]. Retrieving and matching techniques depend on a lexical match between words in users' question and words in documents. In consequence, many unrelated objects will be matched, and related objects will be missed. These techniques yet have no practical solutions to some question types, such as questions that need to be justified [2]. Natural language understanding (NLU) techniques may enable users to pose questions in natural language, and obtain the required information precisely. Natural language understanding is sometimes referred to as an AI-complete problem because natural language understanding requires extensive knowledge about the language and the ability to manipulate it [3]. The most challenging issue in natural language understanding is language is not free from the ambiguity problem.

In general, ambiguity is a pervasive phenomenon in human language [4]. In particular, ambiguity is a critical challenge in extracting semantic of a NLQ posed to a QA system [12]. The ambiguity problem in a natural language can be classified into four types; lexical ambiguity, structural ambiguity, semantic ambiguity, and pragmatic ambiguity [5]. Not all ambiguities can be easily identified and some of them require a deep linguistic analysis. In QA, ambiguity would cause confusion in interpretation of the question, and then affects negatively the accuracy of the overall QA performance. In this paper, we focus on lexical ambiguity resolution for QA. Lexical ambiguity occurs when a word has more than one meaning [3]. For example, given a question*" How can student deposit money into a bank?*", human knows that the bank here refers to a financial institution. Whereas, given a question "*Who is seating on the bank of the river?*", the bank here refers to the sloping land beside the river. But unfortunately, it is very difficult for computers to do the same job. Having more than one meaning for an individual word would lead to matching irrelevant answers and that will decrease the accuracy of retrieving the answers [2].

This paper, proposes a new approach to overcome the lexical ambiguity in the NLQ. All human languages have words that have different meanings in different contexts. To resolve the problem, we must consider the context in which each word and question are posed. Such a process of deciding which of their several meanings is intended in a given context is known as Word Sense Disambiguation (WSD). The proposed new approach integrates context knowledge, and concepts knowledge of interesting domain, into shallow natural language processing (SNLP). Concepts knowledge is modeled using ontology, while context knowledge has been obtained from WordNet. Context knowledge is determined based on neighborhood words in a question.

## 2. Research Background

The most relevant research areas of the proposed approach are QA, natural language processing (NLP), word sense disambiguation (WSD), and ontology. An overview of each area and its related work is presented in the following subsections.

### 2.1 Question Answering (QA)

Question Answering (QA) is a task that combines techniques of information retrieval (IR), template matching, information extraction (IE), and NLP. QA aims to return an exact answer to a natural language question instead of a list of documents. QA system is made up of three modules: processing question processing, documents processing, and answer processing, see Fig 1. In a QA system, answers are normally stored in either structured databases, semi-structured databases or unstructured databases. NLQs are classified into 2 forms [15]; factoid or complex. Factoid questions require simple facts can be found in short text strings, for example, "Where is Microsoft Company located?". Complex questions require first to identify its context, for example, "How can student deposit money into a bank?". Complex questions cannot be answered using the same techniques that apply to factoid questions. QA systems have been evaluated and tracked in several academic workshops such as TREC [42], CLEF [43], and NTCIR [44]. Most of the research work conducted in QA focuses on answering factoid questions [16], [30], and [31]. Answering factoid questions is a simple process that including classification and matching the words in questions with same words in retrieved texts. Whereas, complex questions is a complex process that involves detecting relations among words based on the contextual knowledge [32], [33], and [38]. The need for effective techniques can handle complex questions made QA community to move towards many other new fields (e.g. NLP, knowledge representation (KR), and linguistic). QA systems are classified into two types; closed and open domain. Closed domain systems deal with questions under particular domain [45]. Open domain systems deal with questions about every thing [46].

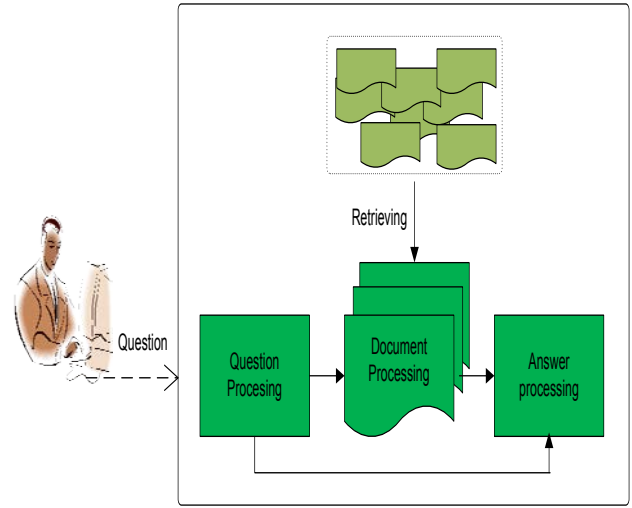

Fig. 1  A basic architecture of a QA system.

### 2.2 Natural Language Processing (NLP)

NLP is a computerized approach concerns with the interaction between computers and human language. The major purpose of NLP is to achieve a human-like language processing that enable computers to understand and a generate language used by humans. Research in NLP has been going since the early 1950s. Machine translation (MT) was the first computer-based application related to natural language. Much research work in NLP have been introduced lately [10], [51], [11], [13], and [14]. However, the goal of NLP is still far from being success [12]. Natural language understanding (NLU) is considered as a subset of NLP. NLU is a system that computes the syntactic and semantic representation of a sentence [17]. Enabling computers to understand the human language is the dream of AI community, therefore several research work attempted to provide those machines with the ability of understanding human natural language [39], [47], and [40].

In NLU there are two important components: syntactic and semantic analysis [18]. Syntactic analysis is a process of assigning a syntactic structure or a parse tree, to a given natural language sentence. Semantic analysis is a process of translating a syntactic structure of a sentence into a semantic representation that is precise and unambiguous representation of the meaning expressed by the sentence. A semantic representation allows a system to perform an appropriate task in its application domain. Current NLP researches in QA field follow two approaches in processing natural questions. Firstly, shallow NLP (SNLP) which concerned with a partial parsing, and do not highly consider the linguistic analyses like in [1], [19], and [41].

The SNLP is performed by several techniques like chunking, keywords matching, pattern matching, and classifying questions and answer types. According to the reference [48], SNLP is processed through several steps: tokenization, part-of-speech tagging (POS), chunking, and shallow paring. Secondly, deep NLP (DNLP) which focus on semantic and contextual processes in the case of analyzing a natural language sentence such as [20], [21], and [22]. DNLP may involve full syntactic parsing, relation detection, and logical inference. The major distinction between the two methods is varying in a degree of considering the semantic issue, and dealing with knowledge of a language, and ways the knowledge is acquired and represented [23]. Although, SNLP approach is considered faster than DNLP, it is not successful to the desired extent. In the same time, a mature and deep syntactic and semantic analysis has no yet performed in QA field [24]. Both SNLP and DNLP employ different methods for computation. They comprise statistical methods [25] and [47], rule-based methods [18], and combination of these [51] and [53].

## 2.3 Word Sense Disambiguation (WSD)

Word sense disambiguation WSD process is required in application such as a QA application [55]. WSD is a process to identify the meaning of a word in a given natural language context. Lexical ambiguity is decided using a dictionary. The most known dictionary used for this task is WordNet, which is described later in this paper. WSD has been recognized as an AI-hard problem [36]. Such a problem in QA would have a significant impact on the accuracy of retrieving answers. WSD concerns with defining the relationships among "word" and "meaning" and "context" [26]. Context is the only means to identify the meaning of an ambiguous word. Therefore, contextual information is required to determine the intended meaning from a set of meanings. The context is determined by two ways [48]: Firstly, relational information which refers to ambiguous word relations, including distance from the target, syntactic relations, selectional p references, orthographic properties, phrasal collocation, semantic categories. Secondly, bag of words which refer to words in the neighborhood in the question without considering their relationships to the ambiguous word.

Research in WSD has a long history. Much of research work in WSD have been conducted [27], [28], and [29]. Reference [3] classified methods applied to these research work into three main approaches: *supervised WSD* which use machine-learning techniques to learn a classifier from labeled training sets. *Unsupervised WSD* which rely on unlabeled corpora, and do not exploit any manually sense-tagged corpus to provide a sense choice for a word in context. *Knowledge-based approach*, this type of approaches depends on external knowledge sources that provide necessary information to associate senses with words. Knowledge sources can vary from corpora of texts, either unlabeled or annotated with word senses, to machine readable dictionaries, thesauri, glossaries, ontologies, etc. Knowledge-based approach has been applied in several works such as [34], [35], [36], [37].

## 2.4 Ontology

Ontology is originally a philosophical term. Ontology is defined as a conceptualization of a domain into a human understandable, machine-readable format consisting of concepts, attributes, relationships, and axioms [56]. A concept represents a set of entities within a domain. Relations describe the interactions among concepts. Relations can be categorized into two main categories: taxonomies that organize concepts into "is-a" and "is-a-member-of" hierarchy, and associative relationships [6]. The associative relationships represent, for example, the functions and processes a concept has or is involved in. Domain ontology also specifies how knowledge is related to linguistic structures such as grammars and lexicons. Therefore, it can be used by NLP to improve expressiveness and accuracy and to resolve the ambiguity of natural language questions [6]. Since, ontology plays an important role in improving efficiency and accuracy of a question answering system. Many researchers employed it in the QA area [6], [50], and [57]. In this research we use ontology to model concepts knowledge of interest domain, in order to resolve lexica ambiguity in NLQ. To the best of our knowledge no a QA system considered concepts knowledge in resolving lexical ambiguity in NLQ.

## 3. Proposed Approach

The proposed approach solves lexical ambiguity in QA by considering two pieces of knowledge: context knowledge, and concepts knowledge. The combination of these knowledge is used to decide the most possible meaning of the word. The Fig 2 illustrates the framework of proposed approach. There are five components, each component described as follows.

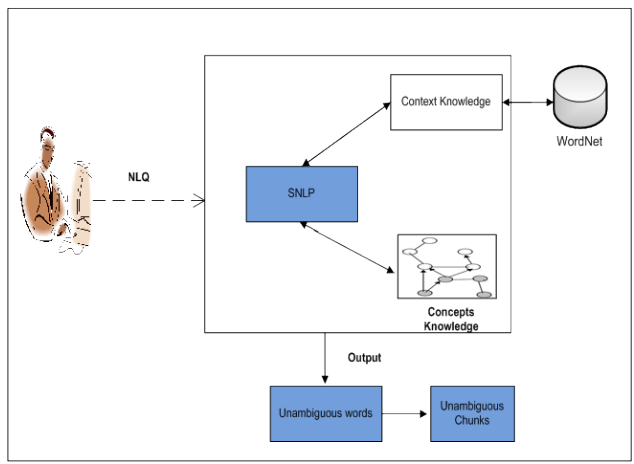

Fig. 2 The framework of a proposed approach.

## 3.1 Natural Language Question (NLQ)

A natural language question can be posed in different ways, e.g. imperative questions and wh-questions. In this paper, we consider Wh- questions. The proposed question processor categorizes the received questions into two structures. Firstly, factoid questions, e.g. "*Who is the chair of the department?*". Secondly, complex questions, e.g. "*How can student contact an advisor in his major?*". Each word in the question will be assigned with its category based on POS process. Then, the question will be classified based on its expected answer type. For example, in the question "*Who is the chair of the department?*", the type of expected answer is PERSON. All unambiguous words which are categorized as Verb or Noun in neighborhood will be used to determine the context of the question.

## 3.2 WordNet

WordNet is a large lexical database of English, developed by Princeton University. The database categorized words into nouns, verbs, adjectives and adverb; each expressing a distinct concept. Nouns, verbs, adjectives and adverbs are grouped into sets of synsets. Synsets are interlinked by means of conceptual-semantic and lexical relations. WordNet is also freely and publicly available on the Internet for download. WordNet's structure makes it a useful tool for computational linguistics and NLP. In this work, WordNet version 3.0 is used to decide if a lexical entry is ambiguous or not, and to provide the context knowledge with the set of possible senses of an ambiguous word.

## 3.3 Context Knowledge

Context knowledge contains a set of lexical with their semantic relations. The set of lexical with its semantics are extracted from the WordNet database manually. All semantics in this work are extracted from the WordNet and combined with a context label. For instance, the word *bank* may have 5 possible meanings as shown in Table I. The proposed processor uses the neighborhood words to determine the context of the question. Unfortunately, as we notice from the Table 1 the same context label can be assigned to deferent senses. Thus, to determine the correct possible meaning, knowledge about lexical meanings and its context are mapped to concepts ontology, which will be described in the next subsection.

Table 1: Context knowledge of the word *bank*

| Sense | Gloss | Context |
|-------|-------|---------|
| #1 | Sloping land | Money, Deposit |
| #2 | Financial institute | River. Lake |
| #3 | container | Money |
| #4 | the funds held by gambling house | Money & Play |
| #5 | a flight maneuver | Transport |
| #6 | a supply or stock held in reserve | Money |

## 3.4 Concepts Knowledge

Concepts knowledge is ontology consists of a set of concepts which are within the domain, and the relationships between the concepts. The ontology also specifies how knowledge is related to linguistic structures such as grammars and lexicons. Fig 3 illustrates a part of ontology of a university domain. The ontology is represented as a graph that consists of nodes (concepts) and edges (relationships). We define the concepts ontology as $O = \{N, E\}$. Where $N$ is a set of nodes which can be represented as $N = \{n_1, n_2, n_3, ..., n_m\}$, where $m$ is a finite integer. R is a set of relations among entities $E = \{e_1, e_2, ..., e_n\}$, where $n$ is a finite integer.

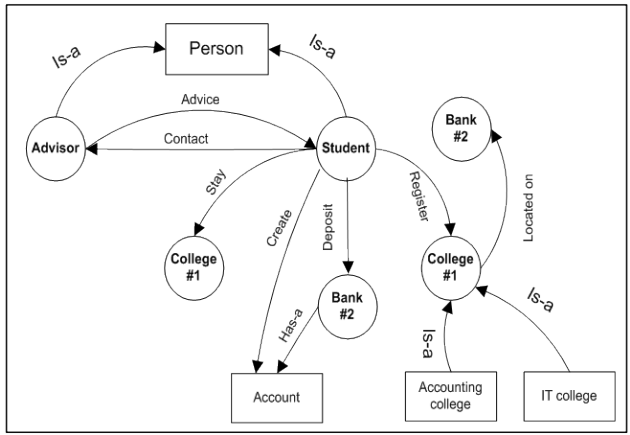

Fig. 3 An example of concepts-domain ontology.

## 3.5 SNLP

The techniques of SNLP that are involved in this work include syntactic processing and semantic processing. Syntactic processing is conducted by implementing a chunker. The purpose of syntactic processing is to recognize syntactic constituents in a sentence. The chunker does not try to analyze a complete question, but only tries to build "chunks" of words from the sentence. The system of chunkers is relatively simple, and efficient.

## 4. Implementation

The steps of the development ased on the given framework is illustrated in Fig 4. There are 6 steps. All steps are equally important. However, Word-sense disambiguation is the core of our NLQ processor in a Q A system. We describe here the all 6 steps:

**Step 1**: This NLQ processor performs the step of part-of-speech tagging (POS). The proposed rule-based tagger reads the question and assigns a class to each word in this question, such as noun, verb, adjective, etc. For this task, we provide the tagger with the necessary linguistic knowledge. For example, given a question "*How can student deposit money into the bank?*", can be tagged as follows:

[How/Wh-Q] [can/Aux] [student/Noun] [deposit/Verb] [money/Noun] [into/ IN] [the/Det] [bank/Noun]

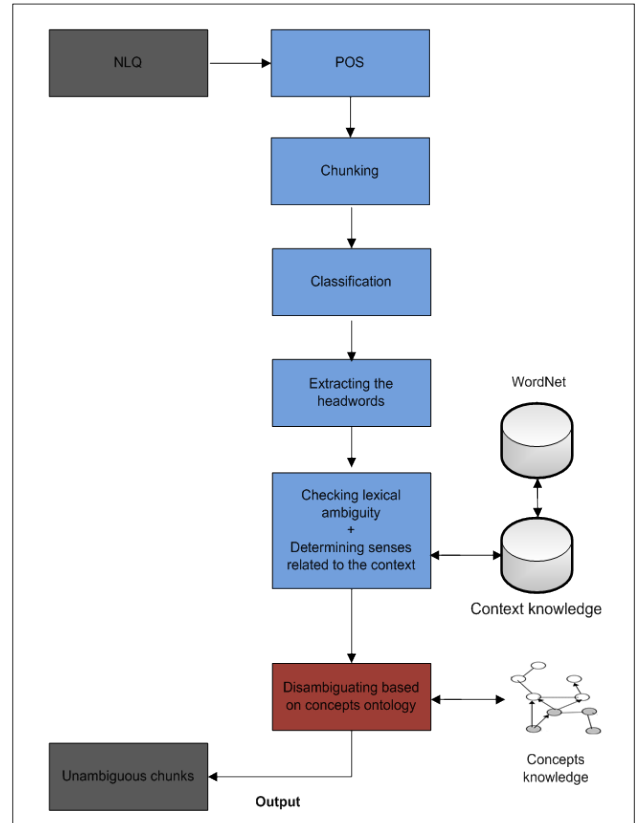

Fig. 4 Steps of the development.

**Step 2**: In this step the question will be identified as constituents (e.g. Noun Phrase (NP), Verb Phrase (VP), and Prepositional Phrase (PP)). For this task, we built a rule-based chunker, which receives a s equence of tagged words, and then divides the question in syntactically correlated segments. For example, given a question "*How can student deposit money into the bank?*", can be divided as follows:

[student/NP] [deposit/VP] [money/NP] [into the bank/PP]

The word "*How*" will be used to classify the type of expected answer, as described in the following step. Question chunking is an intermediate step towards extracting semantic representation of the question which is not the scope of this paper. This work is concerned only with lexical semantic disambiguation. However, the expected output of the proposed processor is unambiguous chunks.

**Step 3**: The task here is to classify the question by its expected answer type. This task will be helpful whether in disambiguation process or in retrieving answers eventually. Knowing the expected type of answer will be helpful more in the questions of factoid type, where some constraints will be imposed on the potential answers. These constraints can be utilized to judge the most correct answer. For example, given a question "*who is the chair of the department?*", this question is expected to be classified into an answer type of PERSON, which is the only candidate answers that are PERSON type need to be considered.

**Step 4**: In this step each word which is assigned as Verb or Noun will be retrieved as headwords. Noun category will be considered as a node (concept) in the ontology, and Verb category as a relation between two nodes. For example, the question "*How can student deposit money into the bank?*" will define *student, money, and bank* words as nodes. Whereas, the word deposit will be defined as a *relation*. This information will be useful in the WSD process in the following steps.

**Step 5**: The output of **step 4** will be examined by the context knowledge to detect the ambiguity of lexical. Ambiguous words (e.g. *bank*) are the main focus of NLQ processor. Unambiguous words will be used to determine the context of the question. For example, the word *money* is considered unambiguous word, so that the processor looks up the context knowledge to find *bank's* senses labeled with *money* context. According to Table 1, there are four senses (#2, #3, #4, and #6) only labeled with *money* context will be considered. The output of this step will serve as an input to the next step.

**Step 6**: In this step the senses (#2, #3, #4, and #6) will be examined by the concepts ontology to disambiguate words. The task here is mapping the concepts based on its relations that extracted from the posed question. To discuss the use of ontology to solve lexical ambiguity let us refer to the Fig 3 in section 3.4. As explained, the word *bank* in the question "*How can student deposit money into the bank*" is ambiguous, and as a result, unrelated answers may be retrieved. In this work, we use the knowledge about concepts of the selected domain. To apply the concept knowledge to the disambiguation process, extracted headwords are mapped to the ontology. When an ambiguous word is detected in a question, the extracted relation and its synonymy will be used in the process of mapping. For example, *deposit* is identified as a relation between the node *bank* and the node *student*. This simple graph is then mapped to the ontology. By doing so, the processor is capable to decide that the sense #2 is the right sense for the word bank in this context.

The output of these processing steps will be unambiguous chunks, which will be helpful in resolving syntactic structure in our future work.

## 5. Conclusions and Future Work

In this paper, a novel approach for resolving lexical ambiguity in natural language questions posed to QA system is proposed. The proposed approach is obtained by combining two pieces of knowledge; context knowledge and ontology of concepts knowledge of interesting domain, into shallow natural language processing (SNLP). The proposed approach is expected to have the ability to overcome the lexical ambiguity problem. The significant contribution of this research work is a new technique for resolving lexical ambiguity in natural language questions posed to a QA system. In the future, we are looking at resolving the syntactic ambiguity in natural language questions.

**Omar Al-Harbi** is a PhD student in Islamic Science University of Malaysia (USIM). He earned his Bachelor degree in computer science from Jerash University, Jordan, and his Master degree in IT from Northern University of Malaysia (University Utara Malaysia UUM). His PhD specializing in Question Answering systems and natural language understanding.

**Shaidah Jusoh** is an associate professor of the Department of Computer Science in the Faculty of Science & Information Technology at Zarqa University, Jordan. She earned degrees of Master in Computer Science and PhD in Engineering System & Computing from University of Guelph, Canada. Her research interests include, an automated text summarization, question-answering system, online social network, data mining, and information extraction.

**Norita Md Norwawi** is associate professor. She is the head of program for computer science at the faculty of science and technology, Universiti Sains Islam Malaysia. She graduated with A BSc in computer science at University of New South Wales, Australia in 1987. In 1994 she obtained her MSc in Computer science from Universiti Kebangsaan Malaysia. She was conferred with a PhD in 2004 from Universiti Utara Malaysia specializing in the area of Artificial Intelligence. She has published her work in peer reviewed journals and international proceedings. Her research interest is in the application of AI techniques especially in information security and assurance such as data mining, multiagent system and visualization.